\documentclass[letterpaper, 10 pt, conference]{ieeeconf}  

\IEEEoverridecommandlockouts                              

\usepackage{graphicx} 
\usepackage{url}
\usepackage{amsmath}
\usepackage{pifont}
\usepackage{stfloats}
\usepackage{tabularx}
\usepackage{threeparttable}
\usepackage{caption}
\usepackage{xcolor}
\usepackage[nocompress]{cite}
\usepackage{hyperref}
\usepackage{tabularx}
\usepackage{siunitx}
\hypersetup{
    colorlinks=true,
    urlcolor=blue,
    citecolor=purple,
    linkcolor=purple 
}
\usepackage{eso-pic}

\definecolor{skyblue}{HTML}{009dff}

\title{\LARGE \bf
NGD-SLAM: Towards Real-Time Dynamic SLAM without GPU}

\author{Yuhao Zhang$^{1,2}$, Mihai Bujanca$^{3}$, Mikel Luj\'an$^{1}$%
\thanks{$^{1}$University of Manchester. $^{2}$University of Cambridge.}%
\thanks{$^{3}$Qualcomm Technologies XR Labs, Austria.}}

\begin{document}

\newcommand{\boldslash}{%
  \leavevmode\rlap{/}\kern0.3pt/%
}
\AddToShipoutPictureFG*{%
  \AtPageUpperLeft{%
    \hspace*{\dimexpr 1in+\oddsidemargin\relax}%
    \raisebox{-0.5in}{%
      \begin{minipage}{\textwidth}
        \footnotesize
        \fontfamily{phv}\selectfont
        \textbf{\copyright~2025 IEEE -- Accepted to IEEE{\boldslash}RSJ International Conference on Intelligent Robots and Systems (IROS) 2025}
      \end{minipage}%
    }%
  }%
}

\maketitle
\thispagestyle{empty}
\pagestyle{empty}

\def\spaceskip{-5pt}


\begin{abstract}

Many existing visual SLAM methods can achieve high localization accuracy in dynamic environments by leveraging deep learning to mask moving objects. However, these methods incur significant computational overhead as the camera tracking needs to wait for the deep neural network to generate mask at each frame, and they typically require GPUs for real-time operation, which restricts their practicality in real-world robotic applications. Therefore, this paper proposes a real-time dynamic SLAM system that runs exclusively on a CPU. Our approach incorporates a mask propagation mechanism that decouples camera tracking and deep learning-based masking for each frame. We also introduce a hybrid tracking strategy that integrates ORB features with optical flow methods, enhancing both robustness and efficiency by selectively allocating computational resources to input frames. Compared to previous methods, our system maintains high localization accuracy in dynamic environments while achieving a tracking frame rate of 60 FPS on a laptop CPU. These results demonstrate the feasibility of utilizing deep learning for dynamic SLAM without GPU support. Since most existing dynamic SLAM systems are not open-source, we make our code publicly available at: \url{https://github.com/yuhaozhang7/NGD-SLAM}

\end{abstract}

\section{Introduction}
\label{sec:introduction}

\begin{figure*}[ht!]
    {
    \setlength{\belowcaptionskip}{\spaceskip}
    \centering
    \includegraphics[width=0.99\textwidth]{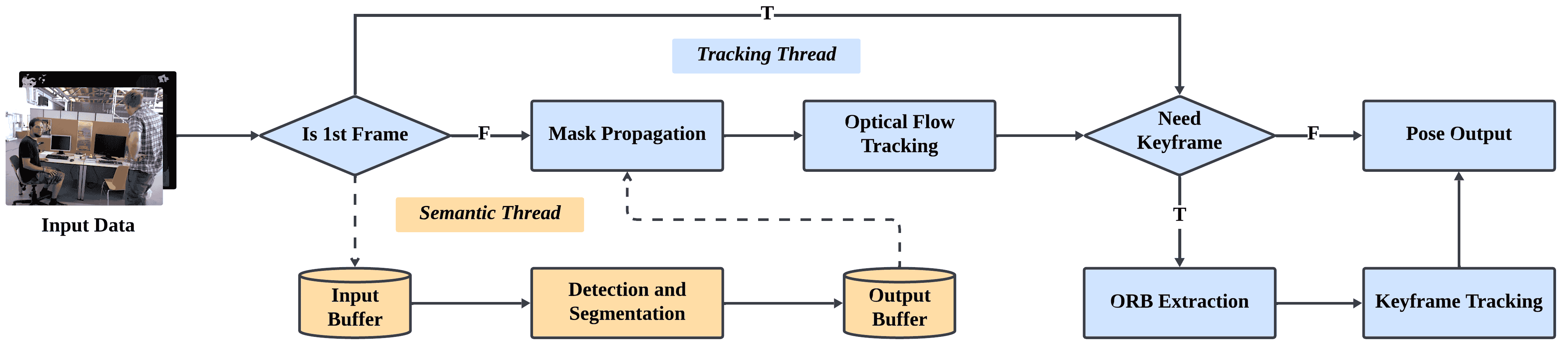}
    \caption{Tracking Pipeline of NGD-SLAM. Based on a mask propagation mechanism, a pixel-wise mask is generated from previous mask to exclude dynamic features. Optical flow is employed to efficiently track the remaining static features across non-keyframes, while ORB feature matching is maintained to establish data association among keyframes.}
    \label{fig:tracking_pipeline}
    }
\end{figure*}

Simultaneous Localization and Mapping (SLAM) is a set of algorithms that leverage sensors' input data to achieve accurate, centimeter-level localization and to build a map of the surroundings simultaneously. It is a foundation for applications like robotics and augmented reality.

One focus in SLAM research is visual SLAM, which refers to using cameras for localization and mapping. Although numerous visual SLAM algorithms \cite{orbslam2, orbslam3, vins, svo} have demonstrated high tracking accuracy across various scenarios, most of them are designed under the assumption of a static environment. These methods typically encounter notable performance degradation in dynamic settings \cite{slamfuse}.

To address this challenge, we propose NGD-SLAM (No GPU Dynamic SLAM), a real-time visual SLAM system for dynamic environments that operates entirely on a CPU.

\textbf{Why CPU Only?}
In visual dynamic SLAM, a common approach is to incorporate deep learning models to mask dynamic objects in input images using semantic information. These methods demonstrate significant improvements compared to approaches that filter dynamic features purely based on geometric constraints \cite{dynamic_slam_survey}. However, these models introduce high computational costs, requiring a GPU to achieve real-time performance on most modern cameras (30 Hz), making them impractical for embedded systems, where weight, power, and heat constraints are critical considerations or GPU may not be available.

\textbf{Why Real-Time?}
One important property of SLAM is its ability to operate online and provide instantaneous pose estimation and map building, both of which are critical for many robotic applications. If a SLAM system operates at a lower frame rate than the sensor rate, it can result in accumulated latency, causing outdated pose estimates; otherwise, it has to drop some input frames to match the system's frame rate, which may degrade performance in high-motion settings due to increased spatial and temporal discrepancies between adjacent frames.
However, many existing dynamic SLAM methods have spent too much computational power on improving tracking accuracy while overlooking the importance of real-time performance \cite{survey_embedded, survey2}.

Therefore, there has been an urgent need to develop robust and efficient dynamic SLAM for low-power devices.
Our system is built upon ORB-SLAM3 framework \cite{orbslam3} and includes an additional semantic thread for dynamic object detection using YOLO \cite{yolo} like many existing methods. The core contribution of this paper is the tracking pipeline shown in Figure \ref{fig:tracking_pipeline}. Rather than using deep learning model less frequently during the tracking process \cite{rdsslam, teteslam, detectslam}, we approach the challenge of improving efficiency by associating previous results with the current frame. Specifically, we introduce a straightforward yet highly effective \textbf{mask propagation mechanism} (Figure \ref{fig:mask_propagation}) that leverages previous segmentation results from semantic thread to predict the current dynamic mask within a few milliseconds. This mechanism enables fully parallel camera tracking and dynamic object segmentation. Considering that frames are not equally important \cite{keyframe, slambook}, we also employ a \textbf{hybrid tracking strategy} that combines the advantages of optical flow and ORB feature tracking to enable better usage of computational resources for each frame and provides much more efficient and robust tracking in dynamic environments.

Through experiments on public datasets, our system demonstrates localization accuracy comparable to that of SOTA methods. When taking real-time performance into consideration, it shows a clear advantage over existing methods, with a tracking frame rate of 60 FPS on a laptop CPU.

The main contributions of this paper are:
\begin{itemize}
    \item An RGB-D SLAM system that not only achieves real-time performance without a GPU but also maintains localization accuracy comparable to state-of-the-art (non-real-time) methods in dynamic environments.
    \item A mask propagation mechanism that efficiently estimates current dynamic object masks based on previous masks, mitigating the high latency caused by the deep learning model.
    \item A hybrid usage of optical flow and ORB features for camera tracking that spend less computation on less important frames (non-keyframes), while maintaining robustness in dynamic environments.
\end{itemize}

\section{Related Work}
\label{sec:related_work}

To address the challenges of dynamic environments, it is crucial to identify pixels or keypoints originating from dynamic objects.
With advancements in deep learning, several methods have been developed for detecting or segmenting dynamic instances across input images using semantic information. Based on ORB-SLAM2 \cite{orbslam2}, DS-SLAM \cite{dsslam} improves localization accuracy in dynamic environments by incorporating SegNet \cite{segnet} for dynamic instances segmentation while maintaining epipolar geometry constraints for moving consistency check. Similarly, DynaSLAM \cite{dynaslam} employs Mask R-CNN \cite{maskrcnn} and multi-view geometry to eliminate dynamic ORB features.
TRS-SLAM \cite{teteslam} steps further towards efficiency by introducing a geometry module that handles unknown moving objects and applying a segmentation model only to keyframes. More recently, CFP-SLAM \cite{cfpslam} achieves high tracking accuracy by proposing a coarse-to-fine static probability approach to enhance localization accuracy by calculating and updating static probabilities as weights for pose optimization. Leveraging the advancements in large segmentation models, USD-SLAM \cite{usd} utilizes SAM \cite{sam} and SegGPT \cite{seggpt} to identify movable objects and applies motion state constraints to handle dynamic instances.

Beyond feature-based methods, some works focus on dense 3D reconstruction and camera tracking in dynamic environments. ReFusion \cite{refusion} uses direct tracking on the TSDF and GPU parallelization to detect dynamics through residuals from pose estimation, while ACEFusion \cite{acefusion} integrates semantic information and dense optical flow to segment dynamic objects and perform both static and dynamic reconstruction on a limited power budget. 

Although incorporating a segmentation model for tracking in dynamic environments gives remarkable improvements in accuracy, algorithms adopting this method typically rely on powerful GPUs to achieve (near) real-time performance due to the computationally intensive nature of deep neural networks. One strategy to improve efficiency is to incorporate a separate semantic thread for the deep learning model. Based on this idea, several approaches have been proposed to parallelize feature extraction and dynamic object identification \cite{dsslam, cfpslam}. However, these methods offer only limited gains in efficiency due to the inherent dependency between the tracking process and the deep neural network. Alternatively, some algorithms apply deep learning only to keyframes \cite{teteslam, rdsslam, detectslam}, which significantly reduces the average tracking time but compromises accuracy.
 
While NGD-SLAM also includes a separate semantic thread for the deep neural network, it incorporates a mask propagation mechanism that allows the tracking thread to operate independently from the semantic thread. Moreover, distinct from the methods that apply dense optical flow or scene flow for differentiating dynamic points from static ones \cite{dsslam, flowfusion, fullfusion, acefusion, rodyn, dense_opticalflow}, NGD-SLAM leverages optical flow to track sparse dynamic and static features, enhancing the system's accuracy and efficiency.

\section{Methodology}
\label{sec:methodology}

In the paper, we refer FPS over 30 as real-time, which is the sensor rate of TUM \cite{tum} and BONN \cite{refusion} datasets.

\subsection{Tracking Pipeline}
Figure \ref{fig:tracking_pipeline} illustrates the tracking pipeline of NGD-SLAM. Initially, the image data of the first frame is forwarded to the semantic thread (Section \ref{sec:semantic_thread}), and this frame is set as a keyframe. ORB features are then extracted and map is initialized as part of the keyframe tracking (Section \ref{sec:keyframe_tracking}). For subsequent frames, the system employs mask propagation mechanism (Section \ref{sec:mask_propagation}) to generate dynamic masks based on previous masks, and uses optical flow to track static keypoints from the last frame and estimates the camera's pose (Section \ref{sec:optical_tracking}). The decision to set a new keyframe is then made. If a new keyframe is required, the system extracts new ORB features from the current frame and discard the features within the mask, and continues with the keyframe tracking using ORB features (Section \ref{sec:keyframe_tracking}).

\subsection{Semantic Thread}
\label{sec:semantic_thread}
NGD-SLAM achieves better run-time performance by adding a semantic thread that detects dynamic objects using the YOLO model \cite{yolo}. This thread works with an input and output buffer, each holding one raw image and the corresponding depth (input) or mask (output). Frames that pass from the tracking thread to the input buffer are processed by the network, then moved to the output buffer to be accessed by the tracking thread. This setup ensures that the semantic thread processes the most recent input frame and the tracking thread does not have to wait for the result from the semantic thread. Instead, it directly takes the mask and corresponding image in the output buffer (without placement) and uses them to predict the current mask. Although there is a temporal mismatch because two threads get invoked at different frequencies, we compensates for this by the mask propagation.

\begin{figure}[t!]
    {
    \setlength{\belowcaptionskip}{\spaceskip}
    \centering
    \includegraphics[width=\columnwidth]{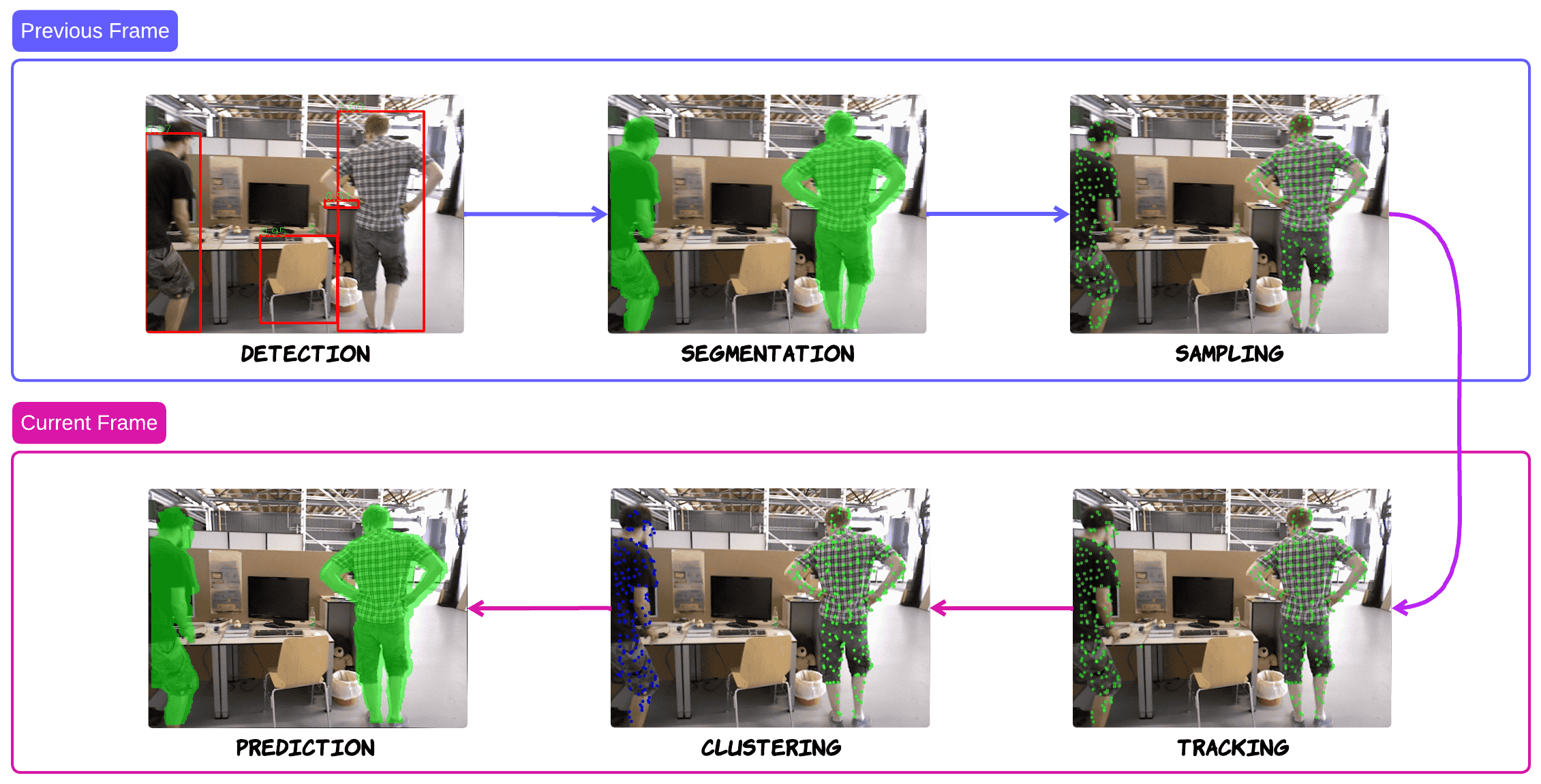}
    \caption{Mask Propagation Pipeline. The detection and segmentation are processed within the semantic thread, while the tracking thread utilizes these results to predict the current mask (the current frame and the previous frame may appear visually similar but are actually different).}
    \label{fig:mask_propagation}
    }
\end{figure}

\subsection{Mask Propagation}
\label{sec:mask_propagation}
One important module of NGD-SLAM is the mask propagation mechanism. Compared to the forward process of a deep neural network, the mask propagation is much faster, taking only a few milliseconds to process one frame on a CPU. It can be expressed as:
\[
M_i = f_\text{prop}(I_i, I_{i-n}, D_{i-n}, M_{i-n}), \quad n > 0,
\]
where $I_{i-n}$, $D_{i-n}$, $M_{i-n}$ represent the raw image, depth image, and mask of the $(i-n)^{th}$ frame, respectively. A smaller $n$ will result in a faster $f_\text{prop}$. 

While mask propagation is also widely studied in video segmentation using deep learning \cite{sam2}, we adopt the classical image processing techniques for NGD-SLAM, as our goal is to achieve supreme efficiency without a GPU, specifically for SLAM applications. Figure \ref{fig:mask_propagation} shows the pipeline of the mask propagation, which consists of six components.

\subsubsection{Detection} 
We employ a relative lightweight YOLO model for the semantic thread, which allows the temporal difference between $I_i$ and $I_{i-n}$ remains small (Figure \ref{fig:mask_propagation}).

\subsubsection{Segmentation} 
Using an object detection model is more efficient than a segmentation network but it only provides bounding boxes for detected objects. To improve this, we use depth information to segment the objects within these boxes, assuming the detection model can accurately locate the object and provide a tight bounding box, such that the detected object occupies most of the box. 
Subsequently, the connected component labeling algorithm \cite{connect_label} is used to exclude masks of objects that share a similar depth but are not part of the main object, and a dilation process further refines the mask before it is passed into the output buffer.

\subsubsection{Sampling} 
Given the masks of identified dynamic objects, dynamic points are extracted within the mask. The image is divided into cells with size of $15 \times 15$. Within each cell, pixels that are masked and meet the FAST keypoint threshold \cite{orbslam3} are identified, and the one with the highest value is selected. If there is no suitable FAST keypoint, a random pixel is chosen within that cell. This grid structure allows for uniform extraction of dynamic keypoints, and the tracking stability is ensured due to the distinctiveness of keypoints and their association with high-gradient regions.

\subsubsection{Tracking} 
The system utilizes the Lucas-Kanade optical flow method to track sampled dynamic keypoints, which assumes that the grayscale value of a keypoint remains constant across different frames \cite{lkopticalflow, slambook}. Mathematically, this is represented as:
\[I(x+dx, y+dy, t+dt) = I(x, y, t)\]
where \(I(x, y, t)\) is the grayscale intensity value of a pixel at coordinates \((x, y)\) with time \(t\). This equation can be further expanded as:
\[I(x+dx, y+dy, t+dt) \approx I(x, y, t) + \frac{\partial I}{\partial x} \cdot dx + \frac{\partial I}{\partial y} \cdot dy + \frac{\partial I}{\partial t} \cdot dt \]
Subsequently, assuming that pixels move uniformly within a small window, an optimization problem can be formulated using the pixel gradients within the window to estimate pixel motion \((dx,dy)\) by minimizing the total grayscale error of matched pixels. For large $n$, tracking will be performed across the intermediate frames between $I_i$ and $I_{i-n}$.

\subsubsection{Clustering} 
For tracked dynamic keypoints in the current frame, the DBSCAN 
algorithm \cite{dbscan} is applied to cluster points that are close to each other while labeling those significantly distant from any cluster as noise. Employing this method allows for the effective distinction between tracked keypoints originating from distinct dynamic objects and filters out the outliers.

\subsubsection{Prediction} 
After deriving the clusters, the system becomes aware of the dynamic entity's location within the current frame. It can then use the depth information of points in each cluster to obtain the corresponding mask with precise shape of the dynamic object, as shown in Figure \ref{fig:mask_propagation}.

\begin{figure}[t!]
    {
    \setlength{\belowcaptionskip}{\spaceskip}
    \centering
    \includegraphics[width=\columnwidth]{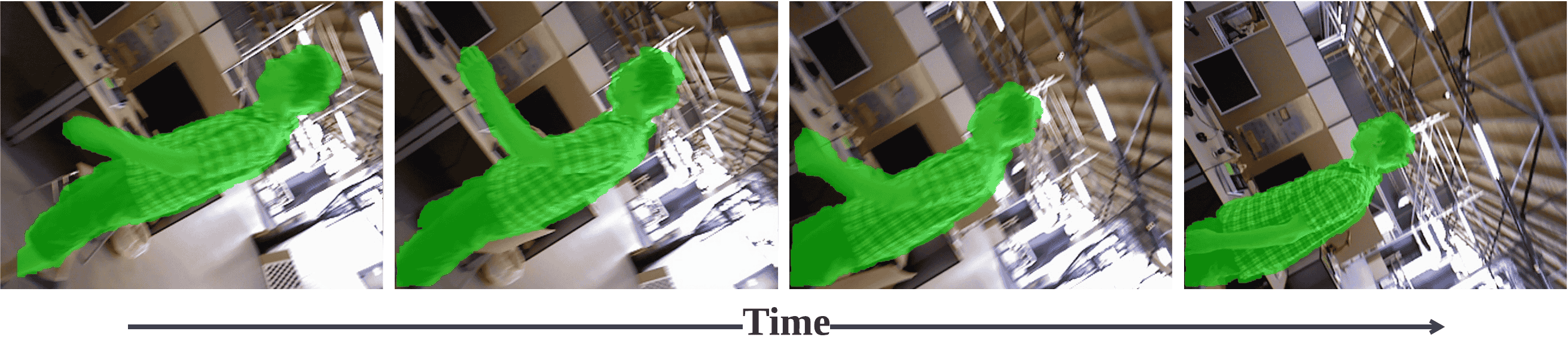}
    \caption{Continuous Mask Propagation. The system continuously tracks a dynamic person over 60 frames by keep propagating the mask from the last frame ($M_{i-1}^p$), when YOLO completely fails with the rotated images.}
    \label{fig:dynamic_tracking}
    }
\end{figure}

In cases where the segmentation model fails to provide a mask, such as when YOLO struggles to detect objects due to camera rotation, the system uses the mask from the last frame to predict the current mask:
\[
M_i^p =
\begin{cases} 
f_\text{prop}(I_i, I_{i-1}, D_{i-1}, M_{i-1}^p), & \text{if buffer empty}, \\
f_\text{prop}(I_i, I_{i-n}, D_{i-n}, M_{i-n}^s), & \text{otherwise.}
\end{cases}
\]
The superscripts denote whether the mask is predicted using the mask propagation mechanism ($p$) or generated by the segmentation model ($s$).
This method allows for continuous tracking of dynamic keypoints until a dynamic object is no longer visible to the camera, even if a deep neural network fails to detect objects. By continually sampling new points from previous mask, rather than tracking the same points across multiple frames, it ensures that dynamic points are up-to-date and enhances tracking robustness (Figure~\ref{fig:dynamic_tracking}).

\subsection{Hybrid Strategy}

\subsubsection{Optical Flow Tracking}
\label{sec:optical_tracking}
In contrast to ORB-SLAM3, which depends on ORB feature matching for every input frame, NGD-SLAM uses the LK optical flow method \cite{lkopticalflow} to track the static keypoints from the last frame. Geometric constraints are formed by associating 3D map points from the last frame to tracked 2D keypoints in the current frame. The pose is then estimated using the EPnP \cite{epnp}, combined with RANSAC \cite{ransac} in an iterative manner.

\begin{figure}[t!]
    {
    \setlength{\belowcaptionskip}{\spaceskip}
    \centering
    \includegraphics[width=\columnwidth]{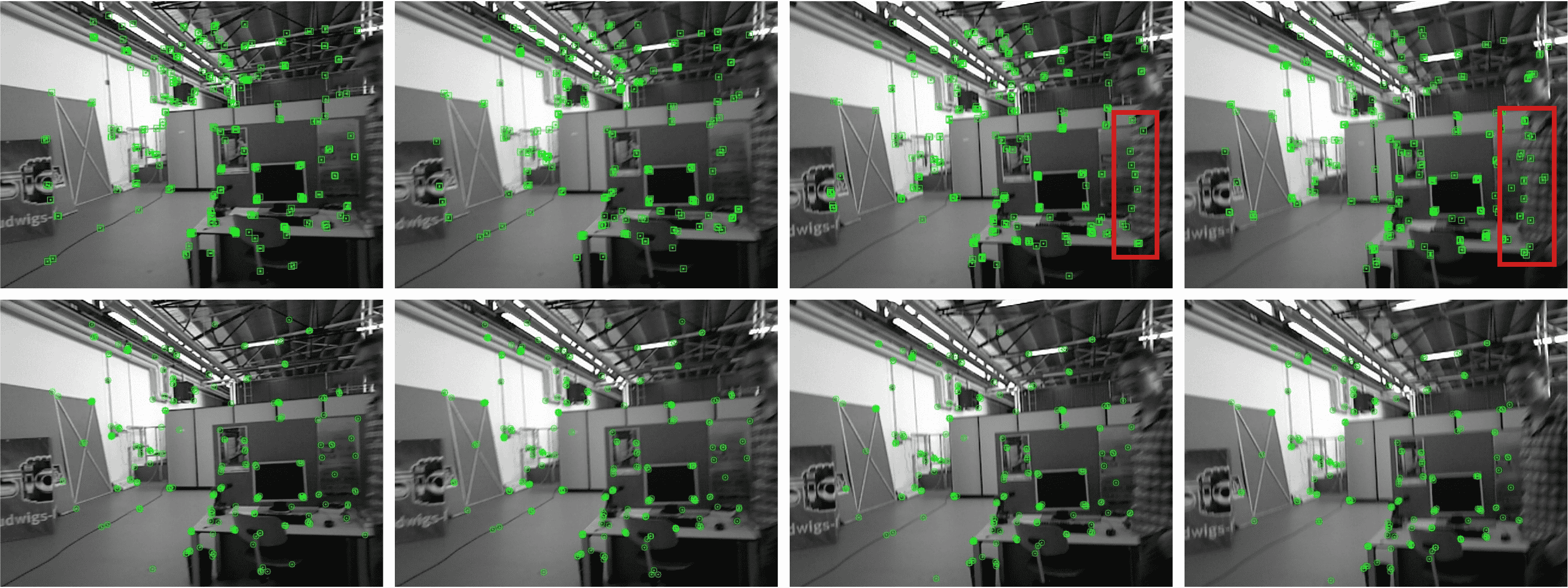}
    \caption{Comparison of ORB Feature Tracking and Optical Flow Tracking. (Top) Extracting new ORB features for each frame and performing matching with map points lead to matches of dynamic points, as shown in the red box. (Bottom) Optical flow tracks same static points across multiple frames, which works well even when mask propagation underperforms due to the newly appeared dynamic person not being detected in previous frames.}
    \label{fig:static_tracking}
    }
\end{figure}

This design is based on two considerations. Firstly, the mask propagation mechanism typically fails when a dynamic object is presented in the current frame but doesn't appear in the frame used for prediction. Thus, extracting new ORB features for each frame and matching with local map points based on the differences in descriptor values can result in mismatches (Figure \ref{fig:static_tracking}). In contrast, optical flow tracking, which only tracks the static keypoints from the last frame based on image gradients and filters out outliers for future tracking, allows for the tracking of the same static points across multiple frames and tends to be more robust in this case.
Secondly, the optical flow tracking skips the computationally expensive ORB feature extraction for non-keyframes, making the tracking process much more efficient.

\subsubsection{ORB Keyframe Tracking}
\label{sec:keyframe_tracking}
ORB features are retained for keyframes to establish connections between them, allowing the system to maintain essential ORB-SLAM3 components such as co-visible graphs and local maps, which are crucial for robust tracking and efficient data retrieval. Keyframes are tracked by first using the pose \( T^\text{init}_t \) to project map points into the current frame to get matches, where \( T^\text{init}_t \) is defined as:
\[
T^\text{init}_t =
\begin{cases} 
T^\text{flow}_t, & \text{if optical flow succeeds}, \\
(T_{t-1} {T_{t-2}}^{-1}) T_{t-1}, & \text{otherwise.}
\end{cases}
\]
Here, \( T^\text{flow}_t \) is the pose of the current frame estimated using optical flow tracking. 
To get the accurate pose, optimization is performed by minimizing the reprojection error:
\[
e = \frac{1}{2} \sum_{i=1}^{n} \left\| u_i - \frac{1}{s_i} KTP_i \right\|^2_2
\]
using the Levenberg-Marquardt method. This optimization process is initialized with \( T^\text{init}_t \) to facilitate faster convergence.
Such keyframe tracking also serves as a backup tracking method when optical flow tracking fails, as ORB features are more robust to illumination changes and motion blur.
Similar to ORB-SLAM3 \cite{orbslam3}, new map points are added after tracking, and the keyframe is passed to the local mapping thread for further optimization.

Compared with many SLAM algorithms that do not distinguish between the computation allocated to keyframes and non-keyframes in frontend tracking, our hybrid usage of ORB features and optical flow to track keypoints allows our system to spend significantly fewer computation on non-keyframes tracking, thereby improving efficiency.

\section{Experiments}
\label{sec:experiments}

\begin{table*}[t!]
\centering
\caption{Comparison of RMSE for ATE (m), RPE translation (m/s), and RPE rotation (°/s) on TUM Highly Dynamic Sequences.}
\renewcommand{\arraystretch}{1.2}
\resizebox{\textwidth}{!}{
\begin{tabular}{|c|ccc|ccc|ccc|ccc|}
\hline
 & \multicolumn{3}{c|}{DynaSLAM} 
 & \multicolumn{3}{c|}{DS-SLAM} 
 & \multicolumn{3}{c|}{RDS-SLAM} 
 & \multicolumn{3}{c|}{TRS-SLAM*} 
 \\
\cline{2-13}
 & ATE (m) & RPE (m/s) & RPE (°/s) & ATE (m) & RPE (m/s) & RPE (°/s) & ATE (m) & RPE (m/s) & RPE (°/s) & ATE (m) & RPE (m/s) & RPE (°/s) \\
\hline
f3/w\_xyz 
& 0.015 & 0.021 & \textcolor{red}{0.452} 
& 0.025 & 0.033 & 0.826 
& 0.057 & 0.042 & 0.922 
& 0.019 & 0.023 & 0.636 \\
f3/w\_rpy 
& 0.036 & 0.045 & 0.902 
& 0.444 & 0.150 & 3.004 
& 0.160 & 0.132 & 13.169 
& 0.037 & 0.047 & 1.058 \\
f3/w\_half 
& 0.027 & 0.028 & 0.737 
& 0.030 & 0.030 & 0.814 
& 0.081 & 0.048 & 1.883 
& 0.029 & 0.042 & 0.965 \\
f3/w\_static 
& 0.007 & \textcolor{red}{0.008} & 0.258 
& 0.008 & 0.010 & 0.269 
& 0.008 & 0.022 & 0.494 
& 0.011 & 0.011 & 0.287 \\
\hline
 & \multicolumn{3}{c|}{CFP-SLAM*} 
 & \multicolumn{3}{c|}{USD-SLAM} 
 & \multicolumn{3}{c|}{DFS-SLAM} 
 & \multicolumn{3}{c|}{Ours*} 
 \\
\cline{2-13}
 & ATE (m) & RPE (m/s) & RPE (°/s) & ATE (m) & RPE (m/s) & RPE (°/s) & ATE (m) & RPE (m/s) & RPE (°/s) & ATE (m) & RPE (m/s) & RPE (°/s) \\
\hline
f3/w\_xyz 
& \textcolor{blue}{0.015} & \textcolor{red}{0.019} & 0.620 
& 0.035 & -- & -- 
& \textcolor{red}{0.013} & -- & -- 
& \textcolor{blue}{0.015} & 0.020 & \textcolor{blue}{0.470} \\
f3/w\_rpy 
& 0.041 & 0.054 & 1.052 
& 0.035 & -- & -- 
& \textcolor{red}{0.027} & -- & -- 
& \textcolor{blue}{0.034} & \textcolor{red}{0.044} & \textcolor{red}{0.889} \\
f3/w\_half 
& \textcolor{blue}{0.023} & 0.027 & 0.785 
& 0.020 & -- & -- 
& \textcolor{red}{0.018} & -- & -- 
& 0.024 & \textcolor{red}{0.025} & \textcolor{red}{0.695} \\
f3/w\_static 
& \textcolor{blue}{0.007} & \textcolor{blue}{0.009} & \textcolor{red}{0.254} 
& -- & -- & -- 
& \textcolor{red}{0.006} & -- & -- 
& \textcolor{blue}{0.007} & \textcolor{blue}{0.009} & 0.262 \\
\hline
\end{tabular}
}
\label{tab:accuracy-cmp-tum}
\begin{minipage}{\linewidth}
{
\vspace{3pt}
\scriptsize
\text{ * indicates methods that are real-time or near real-time, \textcolor{red}{red} indicates the best among all algorithms, and \textcolor{blue}{blue} indicates the best among all (near) real-time algorithms.}
\vspace{\spaceskip}
}
\end{minipage}
\end{table*}

\subsection{Experimental Setup}
Several state-of-the-art and classic feature-based dynamic SLAM algorithms are chosen for comparison, including DynaSLAM \cite{dynaslam}, DS-SLAM \cite{dsslam}, RDS-SLAM \cite{rdsslam}, TRS-SLAM \cite{teteslam}, CFP-SLAM \cite{cfpslam}, USD-SLAM \cite{usd}, and DFS-SLAM \cite{dfs}, as well as dense SLAM methods like ReFusion \cite{refusion} and ACEFusion \cite{acefusion}.

All highly dynamic sequences from the TUM dataset \cite{tum} are utilized, which captures scenarios of two individuals moving around a desk, one wearing a plaid shirt that is rich in texture. We also evaluate the algorithms on sequences of BONN datasets \cite{refusion}, which include situations such as many people walking randomly, a person moving a static object in the scene, several people moving side by side, and the camera continuously tracking a walking person. Two datasets are both captured with a camera operating at 30 Hz.
\begin{table}[t!]
{
\begin{center}
\caption{Comparison of RMSE for ATE (m) on BONN Dataset.}
\label{tab:accuracy-cmp-bonn}
\renewcommand{\arraystretch}{1.2}
\scriptsize
\begin{tabular}{|c|c|c|c|c|c|}
\hline
& DynaSLAM & ReFusion & ACEFusion* & USD & Ours* \\
\hline
crowd & \textcolor{red}{0.016} & 0.204 & \textcolor{red}{0.016} & -- & 0.024 \\
crowd2 & 0.031 & 0.155 & 0.027 & 0.028 & \textcolor{red}{0.025} \\
crowd3 & 0.038 & 0.137 & \textcolor{red}{0.023} & 0.026 & 0.033 \\
mov\_no\_box & 0.232 & 0.071 & 0.070 & -- & \textcolor{red}{0.016} \\
mov\_no\_box2 & 0.039 & 0.179 & \textcolor{red}{0.029} & 0.178 & 0.036 \\
person\_track & 0.061 & 0.289 & 0.070 & \textcolor{red}{0.028} & \textcolor{blue}{0.046} \\
person\_track2 & 0.078 & 0.463 & 0.071 & \textcolor{red}{0.041} & \textcolor{blue}{0.062} \\
synchronous & 0.015 & 0.441 & \textcolor{red}{0.014} & -- & 0.028 \\
synchronous2 & \textcolor{red}{0.009} & 0.022 & 0.010 & -- & \textcolor{red}{0.009} \\
\hline
\end{tabular}%
\end{center}
\vspace{\spaceskip}
}
\end{table}

For metrics, ATE (Absolute Trajectory Error) and RPE (Relative Pose Error) \cite{metric} are chosen as evaluation metrics, with the interval for RPE measurements set to 1 second (30 frames in the case of TUM and BONN). Tracking time per frame is used to measure the efficiency in milliseconds. As some algorithms are not open-sourced and are tested on various devices, a relative fair comparison of efficiency is ensured by displaying a ratio defined as:
\[
\text{Ratio} = \frac{\text{Tracking time of the system}}{ \text{Tracking time of ORB-SLAM2}},
\]
where the information is obtained from their source paper (ORB-SLAM2 is a strong SLAM system for static environments and serves as a common base framework).

All experiments are conducted on a laptop equipped with an AMD Ryzen 7 4800H CPU and an NVIDIA GeForce RTX 2060 GPU, and GPU is not used for our method. 
The presented results are obtained by processing all frames in the sequence, even for methods that are not real-time.

\subsection{Accuracy Analysis}
Table \ref{tab:accuracy-cmp-tum} presents a comparison of our system against various baselines using the TUM dataset, with the minimum error values highlighted. DFS consistently outperforms all four sequences due to its extensive extra computations for handling dynamic objects and static points optimization \cite{dfs}. Our method achieves similar performance with slightly higher error, but when real-time performance is considered, it achieves the highest accuracy overall. While CFP also performs well in near real-time, it is less accurate in the \textit{f3/w\_rpy} sequence as the epipolar constraints it employs are less effective in scenarios involving rotation \cite{cfpslam}.

Table \ref{tab:accuracy-cmp-bonn} shows the results on the BONN dataset. Only Dyna and USD from previous experiment can be used for comparison, as other methods are either not evaluated on this dataset or are not open-sourced, so ReFusion and ACEFusion are added for comparison. Our system outperforms other methods across several sequences. It also has certain ability to handle situations where a walking person is carrying a static object, benefiting from the depth segmentation and the connected component algorithm. This shows the robustness of the system across a wider range of dynamic scenes.

\begin{table*}[ht!]
\centering
{
\caption{Comparison of Tracking Time Ratio (ORB-SLAM2/3 are for Static Environment).}
\label{tab:time-cmp} 
\renewcommand{\arraystretch}{1.2}
\resizebox{\textwidth}{!}
{
\begin{tabular}{|l|c|c|c|c|c|c|c|c|c|c|}
\hline
& ORB-SLAM2 & ORB-SLAM3 & DynaSLAM & DS-SLAM & RDS-SLAM & TRS-SLAM & CFP-SLAM & USD-SLAM & DFS-SLAM & NGD-SLAM \\
\hline
Ratio       & 1.00 & 0.98 & 15.94     & 2.07      & 2.00      & 1.06      & 1.72       & 2.77       & 3.13    & 0.84   \\
Real-Time   & \ding{51} & \ding{51} & \ding{55} & \ding{55} & \ding{55} & \ding{51} & \ding{51} (near)  & \ding{55}  & \ding{55} & \ding{51} \\
Device      & CPU & CPU & CPU + GPU & CPU + GPU & CPU + GPU & CPU + GPU & CPU + GPU & CPU + GPU & CPU + GPU & CPU      \\
\hline
\end{tabular}%
}
\vspace{-5pt}
}
\end{table*}

\begin{table}[t!]
\centering
\scriptsize
\caption{Average Tracking Time per Keyframe (ms).}
\label{tab:time-analysis}
\renewcommand{\arraystretch}{1.2}

\newcolumntype{Y}{>{\centering\arraybackslash}X}

\begin{tabularx}{\columnwidth}{|Y|Y|Y|Y|}
\hline
MP & OF Tracking & ORB Tracking & Total \\
\hline
7.92 & 3.04 & 19.73 & 30.69 \\
\hline
\end{tabularx}

\end{table}

\begin{table}[t!]
{
\centering
\scriptsize 
\caption{ATE RMSE (m) of Different Configurations.}
\renewcommand{\arraystretch}{1.2}

\begin{tabularx}{\columnwidth}{|l|>{\centering\arraybackslash}X|>{\centering\arraybackslash}X|>{\centering\arraybackslash}X|>{\centering\arraybackslash}X|>{\centering\arraybackslash}X|}
\hline
 & ORB3 & +YOLO & +MP & +MP+HS & Improve \\
\hline
f3/w\_xyz 
& 0.390 & 0.014 & 0.014 & 0.015 & 96\% \\
f3/w\_rpy 
& 0.639 & 0.041 & 0.037 & 0.034 & 95\% \\
f3/w\_half 
& 0.431 & 0.023 & 0.028 & 0.024 & 94\% \\
f3/w\_static 
& 0.096 & 0.007 & 0.007 & 0.007 & 93\% \\
\hline
Time / Frame & 20 ms & 54 ms & 28 ms & 17 ms & 15\% \\
\hline
\end{tabularx}

\vspace{3pt}
\parbox{\linewidth}{\scriptsize MP (Mask Propagation), HS (Hybrid Tracking Strategy).}

\vspace{\spaceskip}
\label{tab:ablation}
}
\end{table}

\begin{figure}[t!]
    {
    \setlength{\belowcaptionskip}{\spaceskip}
    \centering
    \includegraphics[width=\columnwidth]{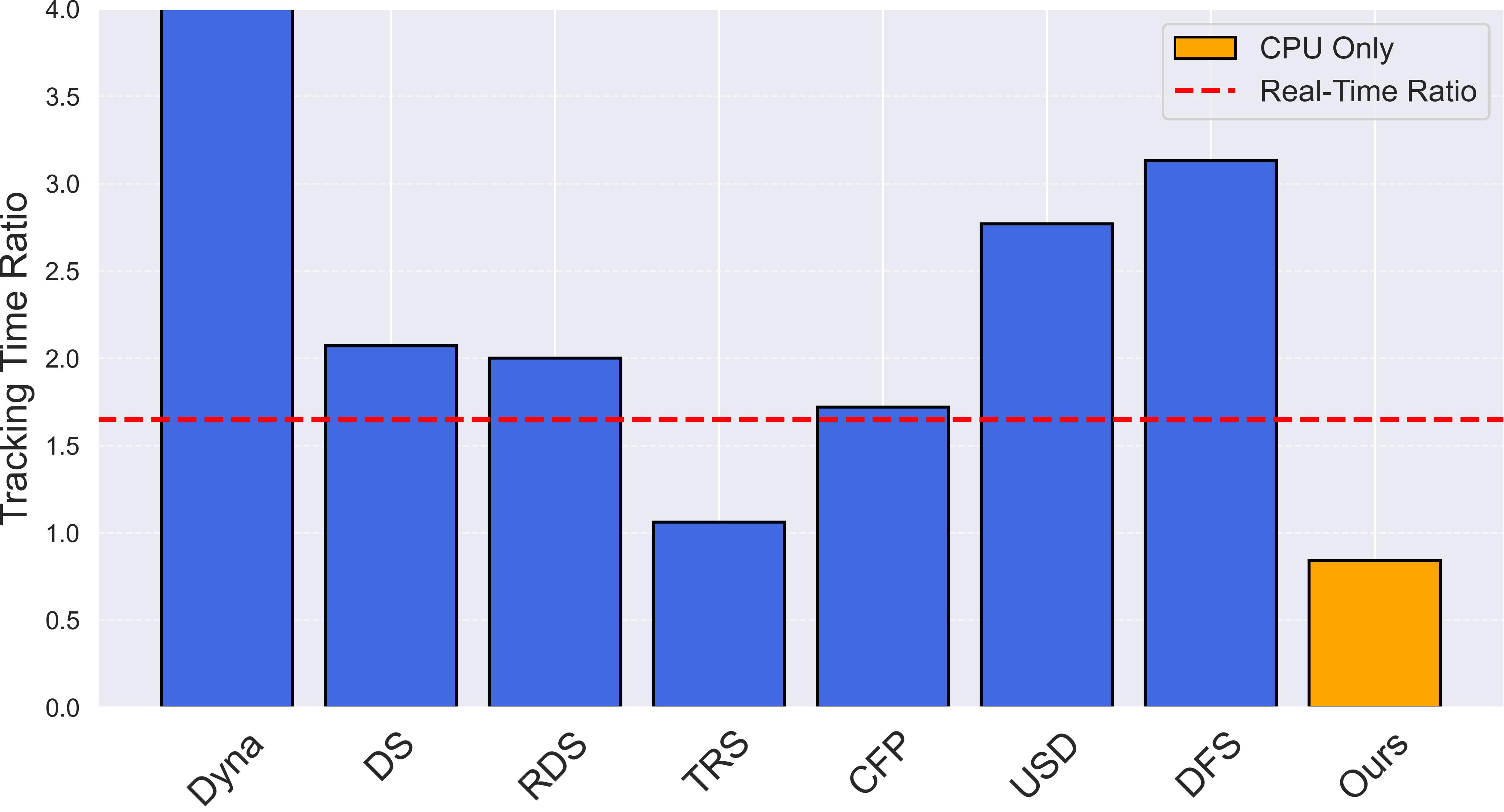}
    \caption{Comparison of the tracking time ratio across different methods.}
    \label{fig:ratio_cmp}
    }
\end{figure}

\subsection{Runtime Analysis}
Table \ref{tab:time-cmp} and Figure \ref{fig:ratio_cmp} presents a comparison of the tracking time ratio across TUM sequences. The average tracking time of ORB-SLAM2 is about 20$\,\mathrm{ms}$ on our experimental device; therefore, a value below 1.65 can be considered real-time (30 FPS in this paper).
While DFS and USD achieve high accuracy, both fall significantly short of real-time performance. Notably, DFS requires more than 3$\times$ the processing time of its base framework ORB-SLAM2, even with GPU. TRS demonstrates clear real-time performance with minimal additional computation for dynamic environments; however, it compromises accuracy by applying deep learning models only to keyframes. In contrast, our system achieves real-time performance with only a CPU while maintaining accuracy comparable to state-of-the-art method. Although some algorithms discussed in the previous section \cite{refusion, acefusion} cannot be fairly compared here, none of them can achieve (near) real-time performance without GPU.

\subsection{Real-Time Analysis}
The average processing time per frame for the system is 16.72$\,\mathrm{ms}$ ($\approx$ 60 FPS), attributed to the efficiency of optical flow tracking in non-keyframes. Since only the tracking of a keyframe involves all components of NGD-SLAM, we also evaluate the average time required for tracking a keyframe using mask propagation, optical flow tracking, and ORB feature tracking, as shown in Table \ref{tab:time-analysis}. It is evident that optical flow tracking is highly efficient, and although mask propagation requires slightly more time, the combined tracking process falls within 33.3$\,\mathrm{ms}$ (30 FPS), which still meets the criteria for real-time performance for the TUM and BONN datasets. Notably, while YOLO detection takes approximately 30 to 40 milliseconds per frame, it is not factored into this tracking time since the tracking and semantic threads operate concurrently at different frequencies.

\subsection{Ablation Study}
Table \ref{tab:ablation} presents an ablation study analyzing the impact of different system components. The base framework, ORB-SLAM3 \cite{orbslam3}, fails to localize in dynamic environments, exhibiting consistently high errors across all sequences.
Integrating a deep learning-based mask significantly reduces errors by filtering out unreliable dynamic keypoints. However, it introduces substantial computational overhead, making real-time tracking infeasible. Mask propagation addresses this by using only previously generated masks rather than waiting for the deep learning model to compute new mask for current frame, enabling real-time tracking on a CPU.
Finally, our full system, incorporating a hybrid tracking strategy, significantly reducing tracking time. By applying fast optical flow tracking rather than ORB feature tracking to non-keyframes, it minimizes the computation allocated to less critical frames.

\subsection{The Importance of Real-Time}

Most existing dynamic SLAM systems use all frames from the datasets during evaluation, even when systems themselves are not real-time. However, such evaluation is insufficient for practical SLAM applications. If a system operates at a lower frame rate than the sensor, latency can accumulate. 
To prevent latency accumulation, which is critical for most SLAM applications that requires instantaneous pose estimation, systems must selectively drop frames, thus matching the input frame rate to the system frame rate. This consideration is frequently overlooked in evaluations of many recent dynamic SLAM algorithms. 

\begin{figure}[t!]
    {
    \centering
    \includegraphics[width=\columnwidth]{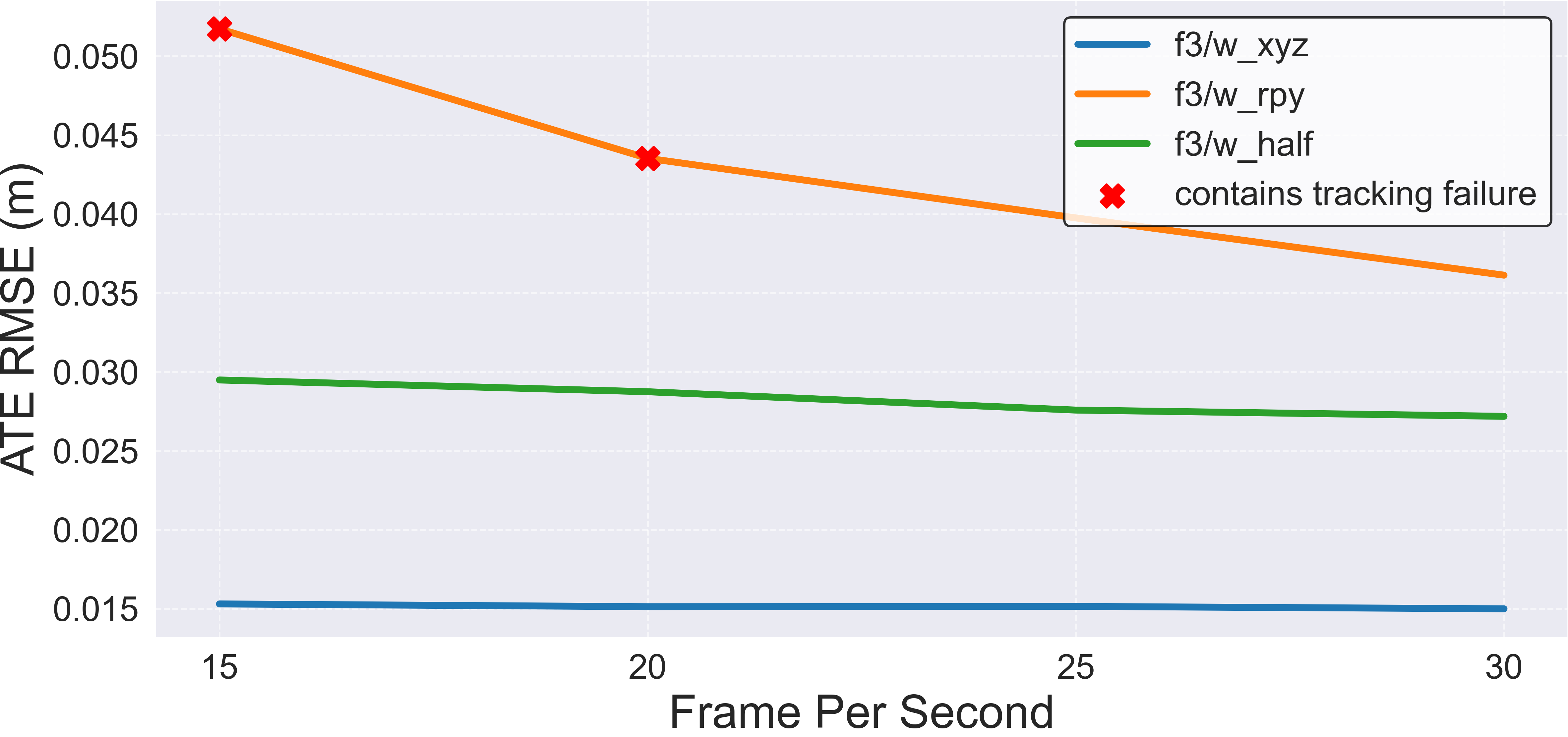}
    \caption{DynaSLAM on TUM Sequences with Reduced Input Frame Rate.}
    \label{fig:fps_cmp}
    \vspace{\spaceskip}
    }
\end{figure}

Since most algorithms are not publicly available, we use DynaSLAM as a reference and evaluated on TUM sequences with reduced input frame rates. As shown in Figure \ref{fig:fps_cmp}, the impact of reduced FPS is minor for sequences involving moderate camera movements. However, in the \textit{f3/w\_rpy} sequence, characterized by rapid camera motion, errors significantly increase and tracking failures occur due to larger temporal gaps between frames at lower frame rates.

Although we were unable to experimentally evaluate SOTA methods such as DFS-SLAM under conditions of dropped frames due to their closed-source nature, it is very likely that similar performance degradation would also occur in non-real-time methods other than DynaSLAM.

\section{Conclusion}
\label{sec:conclusion}

This paper presents a real-time visual SLAM system for dynamic environments that runs on a CPU while maintaining high tracking accuracy. We improve efficiency by decoupling tracking thread from semantic thread at each time step using mask propagation, which can also be easily adapted to other SLAM systems given its generic and modularized design. While this mechanism is based on classical image processing techniques (as we aim to show such a simple method can be highly effective), the boarder concept of mask propagation would allow dynamic SLAM systems to leverage deep learning-based mask propagation for better generalization and robustness, depending on the available resources. In addition, we also propose a hybrid tracking strategy that prioritizes critical frames, reducing computational cost at frontend. Evaluations show that our method achieves accuracy close to SOTA while outperforming real-time methods. Given its high efficiency, we believe it can further benefit from higher-frame-rate cameras, enhancing localization accuracy and high-frequency pose estimation.

\section{Limitation and Future Work}
\label{sec:limitation}

One limitation of our system is that the mask propagation relies on RGB-D inputs for depth information. To adapt it for monocular or stereo SLAM, one could simply use a segmentation model directly rather than a detection model or integrate a learning-based depth estimation model. 

Another limitation is that we do not explicitly handle dynamic instances that do not belong to the classes defined in the detection model or static objects that are movable. However, given the modular nature of mask propagation, it is easy to leverage recent advancements in open-vocabulary segmentation and universal segmentation with motion constriants to identify moveing objects, like USD-SLAM \cite{usd}.

\section*{Acknowledgement}
Huge thanks to Ruiqi Ye (University of Manchester) for the insightful discussions and valuable comments on paper.







\bibliographystyle{ieeetr}
\bibliography{bibliography}

\end{document}